\documentclass{article}
\usepackage{ijcai24}

\usepackage{times}
\usepackage{soul}
\usepackage{url}
\usepackage{xcolor}
\usepackage[hidelinks]{hyperref}
\usepackage[utf8]{inputenc}
\usepackage[small]{caption}
\usepackage{graphicx}
\usepackage{amsmath}
\usepackage{amsthm}
\usepackage{booktabs}
\usepackage{algorithm}
\usepackage{algorithmic}
\usepackage[switch]{lineno}
\usepackage{amssymb}
\usepackage{multirow}
\linenumbers

\title{Learning on Multimodal Graphs: A Survey}

\author{
Ciyuan Peng$^1$
\and
Jiayuan He$^2$\and
Feng Xia$^2$
\affiliations
$^1$Institute of Innovation, Science and Sustainability, Federation University Australia, Ballarat, Australia\\
$^2$School of Computing Technologies, RMIT University, Melbourne, Australia
\emails
ciyuan.p@ieee.org,
jiayuan.he@rmit.edu.au,
f.xia@ieee.org
}

\begin{document}
\nolinenumbers
\maketitle

\begin{abstract}

Multimodal data pervades various domains, including healthcare, social media, and transportation, where multimodal graphs play a pivotal role. Machine learning on multimodal graphs, referred to as multimodal graph learning (MGL), is essential for successful artificial intelligence (AI) applications. The burgeoning research in this field encompasses diverse graph data types and modalities, learning techniques, and application scenarios. This survey paper conducts a comparative analysis of existing works in multimodal graph learning, elucidating how multimodal learning is achieved across different graph types and exploring the characteristics of prevalent learning techniques. Additionally, we delineate significant applications of multimodal graph learning and offer insights into future directions in this domain. Consequently, this paper serves as a foundational resource for researchers seeking to comprehend existing MGL techniques and their applicability across diverse scenarios.

\end{abstract}

\section{Introduction}
Underpinned by significant advancements in artificial intelligence (AI) and machine learning techniques, there is a growing interest in multimodal learning, which involves comprehending and fusing knowledge from diverse modalities, such as texts, images, audios, and various other forms of data~\cite{DBLP:conf/icml/MaZZWFZH23,zheng2023multi}. Compared with unimodal learning which relies on a single data source in one modality, multimodal learning integrates the knowledge from multiple modalities, and hence, allows the development of more effective, accurate, and robust machine learning models, such as large vision models.

As a special type of multimodal learning, multimodal graph learning (MGL) has become an emerging topic, due to the prevalence of multimodal graphs (MGs). Numerous types of multimodal data are present in a graph format, forming MGs where nodes represent entities of heterogeneous types and edges indicate connections amongst them. Notable examples of MGs include healthcare databases of patients' medical records (e.g., lab test results, clinical records, and imaging reports), social networks representing the interactions among internet entities (e.g., online users, organizations, and commercial companies), and interconnected transport networks linking points of interest through different modes of transportation. Hence, developing effective learning techniques over multimodal graphs is regarded as a significant research opportunity, which will provide substantial benefits to downstream applications in various domains, such as large language model~\cite{fu2023mme}, computer vision~\cite{zeng2023multi,saqur2020multimodal}, healthcare~\cite{kim2023heterogeneous,9462380}, and multimedia~\cite{liu2023multimodal,wei2019mmgcn}.

MGL aims to leverage the relational representations of MGs to fully explore the inter- and intra-modal correlations of multimodal data.
A key challenge in MGL involves effectively processing and fusing knowledge extracted from multiple modalities, given the complex graph topology. This is different from other types of multimodal learning where data has a clear and consistent structure. 
In MGL, data fusion must be guided by the graph structure. This is particularly challenging when the graph topology varies across different modalities. Thus, a number of studies have been proposed for MGL, covering a wide range of tasks, including medical tasks (e.g., brain disease detection~\cite{9810283} and drug-drug interaction prediction~\cite{lyu2021mdnn}), 
development of conversation systems (e.g., visual question answering~\cite{DBLP:conf/eacl/HeW23} and conversation understanding~\cite{lian2023gcnet}), information retrieval (e.g., information extraction~\cite{DBLP:conf/acl/LeeLZDPSZSGWALQ23} and image retrieval~\cite{zeng2023multi}), social media analysis (e.g., recommendation~\cite{liu2023multimodal}), and key knowledge graph (KG) tasks (e.g., KG completion~\cite{chen2022hybrid}), etc. These works study different types of MGs, e.g., graphs where data modalities vary within nodes, across nodes, and even across multiple graphs. Meanwhile, they approach the problems of MGL using different techniques, choosing various types of graph learning architectures to achieve effective data fusion in their particular application contexts. The rich diversities of the existing works on MGL call for a comprehensive survey, which systematically summarizes the progress in this domain so far. 

To the best of our knowledge, there is only one survey paper on multimodal graph learning~\cite{ektefaie2023multimodal}, which mainly discusses the potential of leveraging the characteristics of graphs to achieve multimodal learning. It presents how MGs can be constructed in various application scenarios, based on which graph learning technology can be applied. However, it does not discuss the intricacy in the design of multimodal graph learning techniques, i.e., \textbf{how to choose the most appropriate model to process a certain type of MGs}. In this survey, we fill this gap and provide a systematic overview of existing MGL studies in terms of their methodological designs. Particularly, we present an extensive summary outlining the strengths/limitations of popular and elaborately-chosen deep learning architectures for various types of MGs. We summarize vital applications of MGL and list important pointers to useful resources for implementing relevant techniques for these applications. Finally, we discuss the remaining challenges and our insights into future directions. 

\section{Definition of Multimodal Graphs}
In this paper, we define multimodal graphs (MGs) as those graphs that carry data in heterogeneous modalities, such as a combination of visual, textual, and acoustic data~\cite{TianZGMMC22,kim2023heterogeneous}. 
We focus on a popular setting where nodes carry multimodal data while the features of edges are unimodal and reflect the connections of nodes. As such, based on the distribution of different data modalities across nodes, we classify MGs into three types: feature-level MGs, node-level MGs, and graph-level MGs. 

\begin{figure}[h!]
	\centering
	\includegraphics[width=6cm]{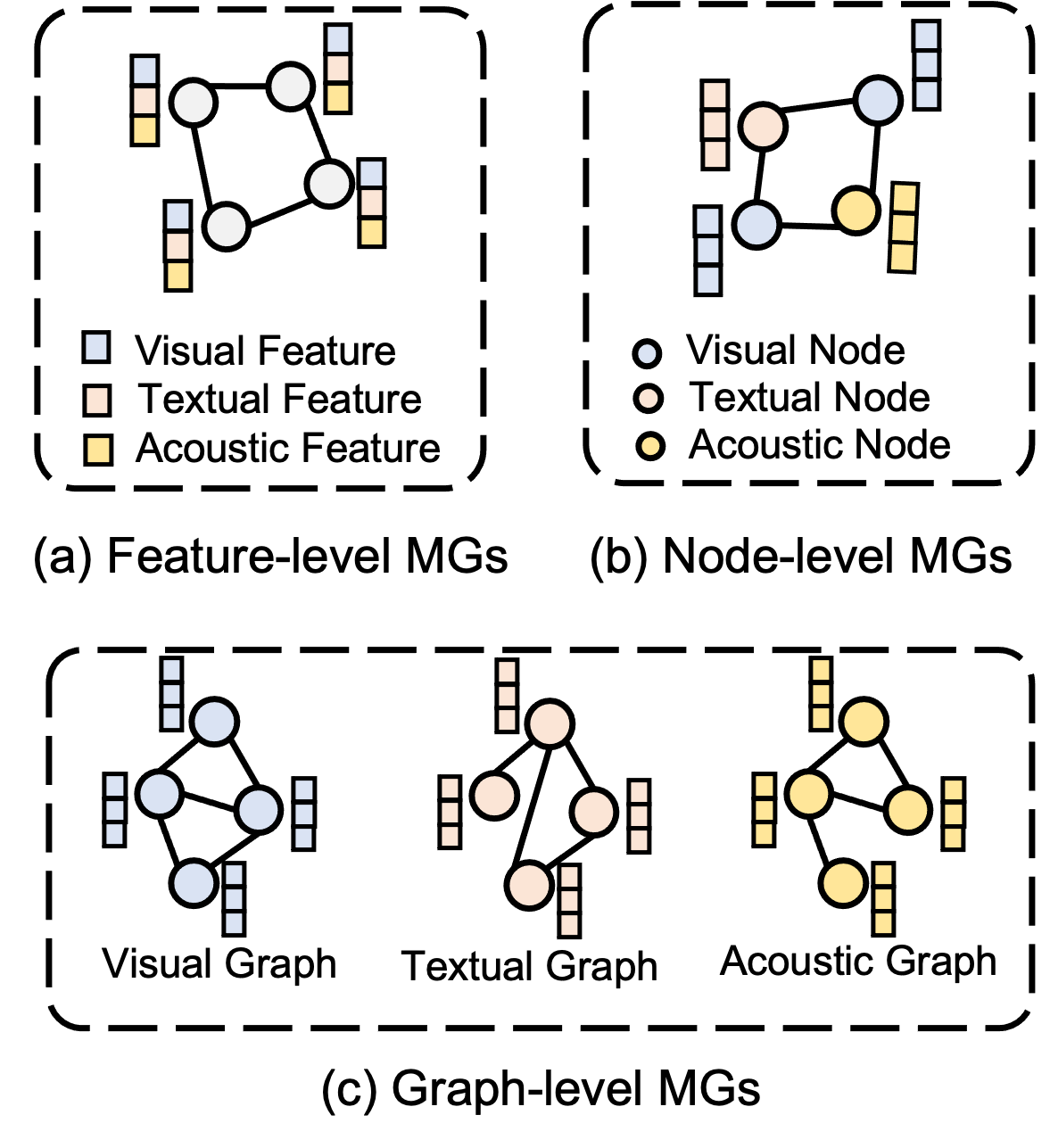} 
	\caption{Three different types of MGs. }
	\label{definition}
\end{figure}

Figure~\ref{definition} illustrates the three types of MGs:
\begin{itemize}
	\item \textbf{Feature-Level MGs:} graphs where each node stores multimodal features. Feature-level MGs are commonly used for a set of interconnected entities, each of which has descriptors of different modalities, e.g., a multimodal conversation system~\cite{lian2023gcnet}. 
	\item \textbf{Node-level MGs:} graphs where each node carries unimodal features while the feature modalities vary across nodes. Node-level MGs are constructed to represent connections amongst entities that are described in different modalities~\cite{hu2023graph}, e.g., a multimodal graph that connects images to their textual keywords.
	\item \textbf{Graph-level MGs:} graphs that contain multiple sub-graphs, each of which stores features of a sole modality. In graph-level MGs, the nodes in sub-graphs usually correspond to the same set of entities, while their connections vary with different modalities~\cite{liu2023multimodal}.
\end{itemize}

\section{A Taxonomy of MGL}
Based on the approaches for processing MGs, we categorize existing MGL works into three predominant types: (1) \textbf{m}ultimodal \textbf{g}raph \textbf{c}onvolution \textbf{n}etwork (MGCN); (2) \textbf{m}ultimodal \textbf{g}raph \textbf{at}tention network (MGAT); and (3) \textbf{m}ultimodal \textbf{g}raph \textbf{c}ontrastive \textbf{l}earning (MGCL). Note that there exist works that use a combination of different learning models for different purposes (e.g., encoders for different modalities). Hence, we categorize all works by their techniques used for the purpose of \textit{data fusion}. Table~\ref{category} presents a list of the representative MGL methods.

\begin{table*}[h]
	\centering
	\renewcommand\arraystretch{0.1}
	\setlength{\tabcolsep}{1.5mm}{
	\begin{tabular}{p{0.14\columnwidth}|l|p{0.075\columnwidth}|p{0.24\columnwidth}|p{0.62\columnwidth}|p{0.45\columnwidth}}
		\toprule
		Category& Method&MG Type&Modality&Task&Description\\
		\midrule
		\multirow{7} {*}{MGCN}
			&\cite{wei2019mmgcn}&F&T+V+A&Recommendation&User-Item Graphs\\
			& \cite{wang2020multimodal}&N&T+V& Content Recognition&KGs\\		
			&\cite{wen2022graph}&N&*&Modality Prediction and Matching&Cell-feature Graphs\\
			&\cite{hu2023graph}& N& V+P&Freezing of Gait (FoG) Detection&FoG graphs\\
		 &\cite{lian2023gcnet}&F&T+V+A&Conversation Understanding&Conversational Interactions\\
		&\cite{9810283}& F &SI+FI+Non-I&Brain Disease Detection&Brain Graphs\\		
		&\cite{zeng2023multi}&N&T+V&Image Retrieval&KGs\\
		\midrule
		
		\multirow{10} {*}{MGAT}
		&\cite{jin2018learning}&N&*& Molecular Optimization&Molecular Graphs\\
		&\cite{chen2020hgmf}&F&T+V+Genetics&Multimodal Fusion&Hypernode Graphs\\
		&\cite{kim2020hypergraph}&G&T+V&Visual Question Answering&Symbolic Graphs\\
		&\cite{tao2020mgat}& G&T+V+A&Recommendation&User-Item Graphs\\
		&\cite{zhang2020multi}& F&T+V& Medical KG Representation Learning&KGs\\
		&\cite{TianZGMMC22}&F&T+V& Recipe Representation Learning&Recipe Graphs\\
		&\cite{chen2022hybrid}& G& T+V&Knowledge Graph Completion&KGs\\
		&\cite{cai2022graph}& G&SI& Brain Age Estimation&Brain Graphs\\
		&\cite{DBLP:conf/eacl/HeW23}& G &T+V& Visual Question Answering&Semantic Graphs\\
		&\cite{000100LDC23}&F&T+V&Analogical Reasoning &KGs\\
		\midrule
		
		\multirow{5} {*}{MGCL}
		&\cite{chen2022multi}& F&T+V &Entity Aligment&KGs\\
		&\cite{lin2022modeling}&G&T+V+A&Sentiment Analysis& Hierarchical Graphs\\		
		&\cite{li2023joyful}&F&T+V+A&Emotion Recognition&Conversational Interactions\\
		&\cite{DBLP:conf/acl/LeeLZDPSZSGWALQ23}& F&T+V&Information Extraction&Document Graphs\\
		&\cite{liu2023multimodal}& G &T+V&Recommendation&User-Item Graphs\\
		\midrule
		
		\multirow{4} {*}{Others}
		&\cite{saqur2020multimodal}& G &T+V&Visual Question Answering&Image/Text Graphs\\
		&\cite{lyu2021mdnn}& N&*&Drug-Drug Interaction Prediction&KGs\\
			&\cite{cao2022otkge}&F&T+V& Knowledge Graph Completion&KGs\\
		&\cite{wang2023hypergraph}&G&SI+FI&Brain Disease Detection&Brain Graphs\\
		\bottomrule
	\end{tabular}}
	\caption{A summary of different MGL methods. ``F" means feature-level MG type, ``N" is node-level MG type and ``G" indicates graph-level MG type. ``T" indicates text, ``V" is vision, ``A" is audio, `P" stands as pressure sequence, ``SI" indicates structural neuroimaging phenotypes, ``FI" is functional neuroimaging phenotypes, and ``Non-I" means non-imaging data. Particularly, ``*" indicates there are various modalities (such as cell, gene expression, protein, molecular and so on) for multimodal biomedical graphs.}
	\label{category}
\end{table*}

\subsection{Multimodal Graph Convolution Network}
Graph convolution network (GCN) is a graph model based on the convolutional neural network that updates node representations through a convolution operation~\cite{wu2019simplifying}. GCN primarily focuses on capturing the relations between nodes and their neighbors, propagating and aggregating information across nodes using adjacency matrices. Recently, to apply GCN over MGs, \textbf{m}ultimodal \textbf{g}raph \textbf{c}onvolution \textbf{n}etwork (MGCN) has been proposed~\cite{wei2019mmgcn,9810283,zeng2023multi}. Notably, due to the advantage of MGCN in exploring the relations amongst nodes, it is highly effective in revealing the correlations between different modalities when applied on node-level MGs~\cite{wang2020multimodal,wen2022graph}. Hence, MGCN shows great potential in handling node-level MGs, compared to the other two types of MGs, in terms of extracting cross-modal relations.

\subsubsection{Cross-modal Relation Modeling for Node-level MGs}
Let $\mathcal{G}=(\mathcal{V},\mathcal{E})$ be a node-level multimodal graph, where $\mathcal{V}$ and $\mathcal{E}$ represent the node and edge set, and $M$ be the set of modalities contained in $\mathcal{G}$. We denote the $i$-th node in $\mathcal{V}$ as $v_i^{m_1}\in \mathcal{V}$ where $m_1\in M$ is its modality. We use $\textbf{X}\in \mathbb{R}^{n\times d}$ to represent the node feature matrix (e.g., $\textbf{x}_i^{m_1}\in \mathbb{R}^d$ represents the feature vector of node $v_i^{m_1}$), where $n$ is the number of nodes and $d$ is the feature dimension. A \textbf{multimodal adjacency matrix} $\textbf{A}\in \mathbb{R}^{n\times n}$ can be obtained. Let $N(v_i^{m_1})$ be the set of neighboring nodes of $v_i^{m_1}$ and $v_j^{m_2}\in N(v_i^{m_1})$ be one neighbour of $v_i^{m_1}$. The adjacency feature $a_{ij}^{(m_1,m_2)}\in \textbf{A}$ can quantify the correlation between modalities $m_1$ and $m_2$. Feeding $\mathcal{G}$ into MGCN, the output of $l$-th layer of MGCN is:
\begin{equation}
	\centering
	\label{eq1}
	\textbf{H}^{(l+1)}=\sigma(\hat{\textbf{D}}^{-\frac{1}{2}}\hat{\textbf{A}}\hat{\textbf{D}}^{-\frac{1}{2}}\textbf{H}^{(l)}\textbf{W}^{(l)}).
\end{equation}
Here, $	\textbf{H}^{(l+1)}$ represents the updated node feature matrix (output of $l$-th layer and input of $(l+1)$-th layer), and $\textbf{H}^{(1)}=\textbf{X}$. $\sigma(\cdot)$is the activation function, $\hat{\textbf{D}}$ indicates the degree matrix of $\mathcal{G}$. $\hat{\textbf{A}}=\textbf{A}+\textbf{I}$, here $\textbf{I}$ is the identity matrix, and $\textbf{W}^{(l)}$ stands as the weight matrix. By updating the weight of $\hat{\textbf{A}}$, MGCN can capture the relations among nodes, and further, extract the cross-modal relations.

In addition to node-level MGs, MGCN has also been applied to other MG types. For instance, GCNet~\cite{lian2023gcnet} leverages a relational MGCN to learn on feature-level MGs for missing modality completion in conversations. \cite{9810283} proposed a multichannel pooling GCN to encode feature-level multimodal brain graphs for brain disease detection. However, MGCN shows more significant power for node-level MGs.


\subsubsection{Characteristics of MGCN}
\textbf{MGCN is effective in learning complex interactions amongst heterogeneous nodes.}~In node-level MGs, neighboring nodes usually have heterogeneous modalities. By processing the multimodal adjacency matrix of a graph, MGCN can propagate the multimodal information residing in one node to its neighbors through convolution kernels, and thus, extract cross-modal relations amongst nodes to achieve effective data fusion~\cite{hu2023graph}. This makes MGCN especially effective when there exist complex interactions amongst the heterogeneous nodes in a graph. For example, MGCN can effectively model the complex interactions in node-level multimodal knowledge graphs, wherein the connections among entities reflect diverse hyper-semantic interactions and cross-modal information (e.g., text-image relations, image-audio relations)~\cite{wang2020multimodal,zeng2023multi}. In such graphs, MGCN explicitly considers the graph topology when aggregating heterogeneous information, and hence, facilitates a more profound understanding of the complicated interactions among multimodal nodes in a graph. Several existing works have demonstrated the superiority of MGCNs in learning effective representations for node-level MGs and achieving enhanced model performance in downstream tasks. 
For instance, GFN~\cite{hu2023graph} utilizes GCN layers to encode multimodal information (e.g., footstep pressure maps and video recordings) in node-level MGs, achieving precise Freezing of Gait (FoG) detection. \cite{wen2022graph} applied MGCN to emphasise the cross-modal relations in multimodal biomedical graphs, performing modality prediction and matching in single cells.

\textbf{MGCN is more effective in short-range information aggregation.} The reach of the convolution operation in MGCNs is limited by its filter size. As such, each node is primarily influenced by its neighboring nodes, and the propagation of information diminishes gradually with the increasing distance. Consequently, works have shown that MGCNs are restricted in capturing long-range information, and therefore, have relatively limited capability to process large-scale MGs. \cite{zhang2020multi} claimed that applying MGCN and its variants (e.g., multimodal R-GCN~\cite{schlichtkrull2018modeling}) to a large multimodal knowledge graph with around sixty thousand entities shows suboptimal performance. Therefore, they utilized an attention-based feature aggregation method to address this limitation of MGCN. Simultaneously, their method indicates the superiority of the graph attention network, which will be discussed in Section~\ref{sec:MGAT}.

\textbf{MGCN is class/label-driven, and does not explicitly guide a model to extract unique features of various modalities.} The strength of MGCN lies in topology-guided feature fusion. However, since the feature fusion is primarily guided by class-driven supervised learning, it does not explicitly guide a model to identify unique characteristics of different modalities in MGs. Consequently, a trained model may not effectively identify the similarities/differences within and across modalities~\cite{liu2023multimodal}. Therefore, several works have proposed to leverage contrastive learning to provide more comprehensive supervision signals, which will be detailed later in Section~\ref{sec:contrastive_learning}. 

\subsection{Multimodal Graph Attention Network}
\label{sec:MGAT}
Graph attention network (GAT) utilizes the attention mechanism to weight the propagated information, dynamically measuring the importance of nodes~\cite{velickovic2018graph}. Compared to GCN, GAT can flexibly learn the different attention weights of node neighbors and more effectively capture long-range information~\cite{wang2019kgat}. Especially, when dealing with large-scale graphs, GAT can adaptively focus on important nodes by assigning them high attention weights, therefore achieving efficient learning. There are various GAT types according to different attention mechanisms, such as self-attention. \textbf{M}ultimodal \textbf{g}raph \textbf{at}tention network (MGAT) extends GAT to the multimodal settings, integrating information from different modalities~\cite{DBLP:conf/eacl/HeW23,cai2022graph}. 

\subsubsection{Attention-based Multimodal Information Fusion}
MGAT implements multimodal information fusion by assigning different attention weights to nodes~\cite{TianZGMMC22}. For a given MG with $|\mathcal{V}|$ nodes, the latent representation of node $v_i$ is calculated as:
\begin{equation}
	\centering
	\label{eq2}
	\textbf{h}_i=\delta(\sum_{v_j\in N(v_i)}\alpha_{ij}\textbf{W}\textbf{h}_j),
\end{equation}
where  $\delta(\cdot)$ is the activation function (e.g., $\texttt{LeakyReLU}(\cdot)$), $N(v_i)$ represents the neighbour set of node $v_i$, and $\textbf{W}$ is a learnable weight matrix. $\alpha_{ij}$ stands as the attention weight, which is calculated as:
\begin{equation}
	\centering
	\label{eq3}
	\alpha_{ij}=\frac{exp(e_{ij})}{\sum_{v_k\in N_{(v_i)}}exp(e_{ik})}.
\end{equation}
Here, $e_{ij}$ is the relation weight of nodes $v_i$ and $v_j$. Equation~\ref{eq2} is a standard formulation of MGAT, and can be optimized according to the setting of attention mechanisms. For node-level and feature-level MGs, MGAT integrates node features to obtain multimodal graph representations, denoted as $\textbf{H}=[\textbf{h}_1,\textbf{h}_2,...,\textbf{h}_{|\mathcal{V}|}]^{\mathsf{T}}$. Here, for feature-level MGs, multimodal features are expressed in each node vector. For example, the attributes of $\textbf{h}_i$ are multimodal. For node-level MGs, the attributes of each node are unimodal, while the modalities across nodes are different, denoted as $\textbf{H}=[\textbf{h}_1^{m_1},\textbf{h}_2^{m_2},...,\textbf{h}_{|\mathcal{V}|}^{m_{|\mathcal{V}|}}]^{\mathsf{T}}$, here $m_i$ indicates the modality of node $v_i$. If nodes $v_i$ and $v_j$ have the same modality, their corresponding $m_i$ and $m_j$ will be the same. In particular, for a node-level MG, because each node is assigned different attention weights, the overall contribution of each modality to the final graph representation will be different. Therefore, a trade-off parameter has been suggested to balance the contribution of multiple modalities~\cite{DBLP:conf/eacl/HeW23}.

Different from node-level and feature-level MGs, for graph-level MGs, MGAT first separately learns on each unimodal graph, and then utilizes a \textbf{co-attention mapping} to integrate graphs from different modalities, obtaining multimodal graph representations~\cite{kim2020hypergraph,chen2022hybrid}. Formally, given a set of graph-level MGs containing $M$ unimodal graphs, the representations of all the unimodal graphs can be gained by Equation~\ref{eq2}, denoted as $\{\textbf{H}^1,\textbf{H}^2,...,\textbf{H}^M\}$. Then, a co-attention mapping function $f(\cdot)$ is applied to obtain a multimodal graph representation. In literature, various methods have been proposed to construct the co-attention mapping function, such as semantic similarity measuring~\cite{kim2020hypergraph} and gated attention aggregation~\cite{tao2020mgat}.


\subsubsection{Characteristics of MGAT}

\textbf{MGAT may provide higher training efficiency.}~MGAT captures node-level correlations via an attention mechanism, dynamically assigning attention weights to various nodes in MGs. Hence, it can potentially lead to more efficient training with reduced computational cost, especially when nodes have varying degrees of importance. As demonstrated by prior studies, attention-based information fusion models have fast training speed and superior performance when dealing with large heterogeneous graphs~\cite{yang2023simple}. This indicates the ability of MGAT to perform efficient model training.

\textbf{MGAT can capture long-range inter-modal dependencies.} MGAT excels in aggregating multimodal information globally over MGs. This is because the information fusion in MGAT is performed through an attention mechanism, which has long-range reach. Many works have proven the attention mechanism empowers each node with a global perception, allowing it to consider information from all other nodes in the heterogeneous graph rather than being confined to a local neighborhood~\cite{wang2019heterogeneous}. As such, MGAT is particularly advantageous when handling large-scale MGs.

\textbf{MGAT may form biased multimodal representations for node-level MGs.}~MGAT treats nodes differently by assigning them different importance scores. Hence, in node-level MGs, nodes from each modality will have different overall impacts on the final learned multimodal representations. This may lead the model to mistakenly consider certain modalities as less important, subsequently ignoring the information from these modalities and resulting in biased multimodal representations. Referring to one existing work~\cite{DBLP:conf/eacl/HeW23}, which introduces a trainable bias to guide the information flow, a potential solution is introducing a trade-off parameter to implement a flexible attention mechanism and balance the contribution of each modality.

\textbf{MGAT is less robust when learning on noise-prone modalities.} Due to the noise-sensitive nature of the attention mechanism, MGAT is susceptible to irrelevant information from some noise-prone modalities (such as images and audios). \cite{chen2022hybrid} demonstrated that ignoring the noise accompanying irrelevant information may lead to modality contradiction. Therefore, the weakness of the MGAT in handling noise may result in biased multimodal representations, especially when dealing with noise-prone modalities.

\subsection{Multimodal Graph Contrastive Learning}
\label{sec:contrastive_learning}
Graph contrastive learning (GCL) focuses on learning distinct node representations by comparing positive samples (similar nodes) and negative samples (dissimilar nodes)~\cite{you2020graph}. Conceptually, GCL aims to maximize the feature consistency of similar nodes and minimize that of dissimilar nodes under different augmented graph views. Benefiting from the efficacy of GCL in enhancing the semantics of node representations and distinguishing different nodes, \textbf{M}ultimodal \textbf{g}raph \textbf{c}ontrastive \textbf{l}earning (MGCL) applies GCL to MGs~\cite{li2023joyful,lin2022modeling}. 

\subsubsection{Inter-modal and Intra-modal Difference Extraction}
Unlike MGCN and MGAT, MGCL explicitly emphasizes on extracting inter- and intra-modal similarities/differences from MGs. It generates different graph views from different perspectives (e.g., modalities), and then applies the contrastive learning strategy to those graph views to learn distinguishable node representations. Given an MG and the latent representation of an anchor sample $\textbf{x}_i$, contrastive learning aims to minimize the following loss:
\begin{equation}
	\label{eq4}
	\mathcal{L} =-\log \frac{exp(h(\textbf{x}_{i},\textbf{x}_{i}^{pos})/\tau)}{\mathop \Sigma_{k=1}^{K}
		exp(h(\textbf{x}_{i},\textbf{x}_{k})/\tau)},
\end{equation}
Here, $h(\cdot)$ is a score function that measures the similarity between two sample representations. $\textbf{x}_{i}^{pos}$ is the representation of a positive sample (similar node) w.r.t. the anchor node, and $\textbf{x}_{k}$ indicates a negative sample. $K$ and $\tau$ represent the number of negative samples and the temperature parameter, respectively.

For graph-level MGs, MGCL can generate different graph views according to modalities. Given a graph-level MG with $M$ unimodal sub-graphs, each unimodal graph is encoded by a graph neural network separately, resulting in $M$ unimodal graph views $\{\mathcal{G}^1,\mathcal{G}^2,...,\mathcal{G}^M\}$. Here, a unimodal graph view 
is denoted as $\mathcal{G}^{m_1}=(\textbf{X}^{m_1},\textbf{A}^{m_1})$, where $\textbf{X}^{m_1}$ indicates the node representation of graph in modality $m_1$, and $\textbf{A}^{m_1}$ is the adjacency matrix. Then, Equation~\ref{eq4} can be applied to capture the similarity and dissimilarity of nodes across different modalities, thereby capturing inter-modal differences and similarities~\cite{liu2023multimodal}. On the other hand, for feature-level and node-level MGs, MGCL obtains two distinct MG views by performing some graph augmentation methods, such as different transformations of MG structure~\cite{chen2022multi,DBLP:conf/acl/LeeLZDPSZSGWALQ23}. Then, Equation~\ref{eq4} is leveraged to distinguish node representations and capture inter- and intra-modal differences. Notably, the trade-off of inter-modal and intra-model difference extraction can be reflected by the design of positive and negative samples~\cite{lin2022modeling}. For example, setting nodes from different modalities as negative samples can capture inter-modal differences, while setting negative samples within the same modality can explore intra-modal differences.


\subsubsection{Characteristics of MGCL}
\textbf{MGCL allows a model to learn correlations/dissimilarities amongst various modalities.}~As a self-supervised learning strategy, MGCL does not require any manual labels from training data. This leads to an obvious advantage of MGCL: lifting the requirement of data annotation and the vast amount of unlabelled data can be utilized for training. More importantly, MGCL generates effective supervision signals through contrasting training samples, which explicitly guides a model to learn the unique characteristics of each modality and the inter-modal correlations. Consequently, this allows a model to link highly correlated modalities and distinguish differences across modalities, forming more comprehensive multimodal representations. Several studies~\cite{chen2022multi,li2023joyful} have shown that incorporating contrastive loss into the traditional class-driven loss function can improve the model performance significantly.

\textbf{Extending MGCL to graph-level MGs with more than two modalities remains a challenge.}~Although MGCL can capture inter-modal similarities and differences when applying it to graph-level MGs, the contrastive loss function is usually implemented on two graph views~\cite{lin2022modeling}. How to efficiently extend MGCL to graph-level MGs with more than two modalities is still an unexplored question in this area. When there are more than two modalities, Equation~\ref{eq4} is no longer suitable. Thus, an advanced loss function is required to capture the relations between \textit{each} pair of modalities, which may make the training process more complex. 

\textbf{The effectiveness of MGCL depends on the design of positive and negative samples.} A critical component in MGCL is to define positive/negative samples in Equation~\ref{eq4}. These samples have significant impact on the quality of learning results. Particularly, the design of positive/negative samples can influence inter- and intra-model difference extraction in MGs. As such, users of MGCL need to carefully define positive/negative samples given their application scenarios, e.g., the modalities involved and the graph topology~\cite{DBLP:conf/acl/LeeLZDPSZSGWALQ23}.

\subsection{Other Methods}
In addition to the aforementioned three representative categories of methods, there are also other methods for processing MGs. We summarize them in this section. 
For example, OTKGE~\cite{cao2022otkge}, a multimodal knowledge graph embedding method, leverages multimodal knowledge (visual and textual) for the representation of entities and relations. \cite{lyu2021mdnn} proposed an MDNN model, in which they employed a graph neural network module to learn the multimodal drug knowledge graph, for drug-drug interaction prediction. 
\cite{saqur2020multimodal} designed a graph parser and a graph matcher. In the graph parser, they respectively extracted two multimodal graphs, then applied a two-layer graph isomorphism network to learn two graphs and fuse their representations to get the multimodal representation in the graph matcher. Moreover, \cite{wang2023hypergraph} presented HMGD for brain disorder diagnosis. They proposed a multimodal phenotypic graph diffusion method to integrate multimodal brain graphs.

\section{MG Libraries and Applications}

\subsection{Multimodal Knowledge Graphs}
Multimodal knowledge graphs (MKGs), one of the most representative MGs, have attracted extensive attention in related domains such as natural language processing and computer vision, e.g., developing multimodal large language models (MLLMs)~\cite{fu2023mme,cui2024survey} for construction of MKGs. MKGs organize multimodal facts with entities and their relations from multiple sources, such as text, image, and video, comprehensively describing various knowledge with significantly improved quality~\cite{chen2022hybrid,DBLP:conf/emnlp/ZhaoCWZZZJ22}. As groundbreaking advancements in information expression, MKGs have been applied in various downstream applications, including dialogue systems, recommender systems, and information retrieval systems~\cite{yang2021unimf}. These applications leverage MKGs' ability to integrate diverse modalities to enhance their performance and effectiveness. Recently, MKG-related tasks, such as multimodal named entity recognition (MNER), multimodal link prediction (MLP), multimodal relation extraction (MRE) and multimodal knowledge reasoning (MKR), have emerged as the primary tasks of knowledge graph embeddings~\cite{chen2022hybrid}. Moreover, many domain-specific MKGs have been proposed to achieve multimodal relation modeling in different areas~\cite{cao2022otkge}. For example, medical MKGs have shown great potential in precise drug event prediction by integrating multimodal drug-drug interactions~\cite{krix2023multigml}.

\textbf{Libraries.}~Publicly available libraries are essential to facilitate the research and applications of MKGs. Three common MKG libraries are WN18-IMG~\cite{chen2022hybrid}, WN9-IMG~\cite{DBLP:conf/ijcai/XieLLS17} and FB15K-237-IMG~\cite{toutanova2015representing}, in which each entity has $10$ images. Some other MKG datasets are also widely used, such as MMDialKB~\cite{yang2021unimf} building upon human-human dialogues, MARS~\cite{000100LDC23} for multimodal analogy reasoning task, and disease-disease KG~\cite{lin2023multimodal} for disease relation extraction.

\subsection{Multimodal Biomedical Graphs}
Multimodal biomedical graphs (MBioGs) serve as powerful tools for organizing diverse biomedical information across different dimensions. They store comprehensive knowledge of biological systems, providing a holistic view that encompasses both molecular and cellular levels. At the biological molecular level, \textit{multimodal molecular graphs (MMGs)} transcend traditional atom-centric representations, capturing the stereochemical structures of molecules along with their diverse molecular paraphrases~\cite{guan2021regio,mercado2021graph}. MMGs can depict intra- and inter-molecular properties from multiple scales, modeling molecule reactions and applied to artificial intelligence-informed medical analysis, such as drug discovery and molecular optimization~\cite{jin2018learning,DBLP:conf/iclr/FuGXYCS22}. 

At the biological cell level, \textit{multimodal single-cell graphs (MSGs)} leverage multi-scale interactions within single cells, such as protein-protein interactions and gene expressions, to measure comprehensive biological information~\cite{gainza2020deciphering}. Learning on MSGs provides feature representations of key biomedical entities (e.g., proteins, drugs, and genes), capturing high-order relations and significantly benefiting single-cell analysis~\cite{wen2022graph}. MBioGs contribute to advancing our understanding of complex biological systems by integrating multimodal data sources. 

\textbf{Libraries.}~Some open biomedical libraries provide available datasets for MBioG analysis. ZINC dataset~\cite{irwin2012zinc}, containing a vast amount of small molecular compounds, is an open database providing resources for research in bioinformatics, drug discovery and computational chemistry. More recently, a multimodal single-cell dataset~\cite{luecken2021sandbox} has been released for multimodal single-cell integration analysis. In addition, various other biomedical corpora (e.g., GENIA corpus~\cite{kim2003genia} and BioNLP Shared Task~\cite{nedellec2013overview}) can be utilized to construct MBioGs.

\subsection{Multimodal Brain Graphs}
Various neuroimaging techniques, including electrogastrography (EGG), magnetic resonance imaging (MRI), positron emission tomography (PET), and diffusion tensor imaging (DTI), are developed to collect multimodal neuroimaging data for brain health analysis~\cite{9810283}. In particular, to explore the complex information interaction mechanism among brain regions, multimodal brain graphs (MBrainGs) are proposed to integrate multimodal neuroimaging data and depict various brain structural and functional connectivities~\cite{9462380}. Many studies have shown the superiority of MBrainGs in capturing multi-dimensional relations of brain regions~\cite{9933896}. 

According to the characteristics of neuroimaging data, MBrainGs can be categorized into several different types. Firstly, \textit{multimodal brain functional graphs (Mbrain-FGs)} are constructed by exploring brain functional connectivities based on multimodal functional neuroimaging data (e.g., Magnetoencephalography (MEG) and EEG)~\cite{chen2023orthogonal}. Since they capture diverse functional information, Mbrain-FGs can be employed for brain activity analysis. Secondly, \textit{multimodal brain structural graphs (Mbrain-SGs)} fuse structural information from multimodal structural neuroimaging data (e.g., structural MRI and DTI) to model brain structural connectivities~\cite{cai2022graph}. Thirdly, \textit{multimodal brain structural-functional graphs (Mbrain-SFGs)} integrate both structural and functional data to achieve joint representations of brain structure and function, showing great potential in advancing various applications, especially precise brain disorder diagnosis~\cite{wang2023hypergraph}. In addition, some works also embed features extracted from non-imaging data (e.g., emographic data) into MBrainGs to enhance the graph~\cite{kim2023heterogeneous}. Learning on MBrainGs provides researchers with a more comprehensive understanding of the brain structure and function, facilitating the in-depth exploration of multi-dimensional relations within the brain.

\textbf{Libraries.}~Publicly available brain health datasets provide essential support for research on MBrainGs. ADNI~\cite{wang2023hypergraph} and OASIS-3~\cite{kim2023heterogeneous} datasets collect the multimodal neuroimaging data of subjects from different age groups for studies on Alzheimer's Disease. ABIDE~\cite{kim2023heterogeneous} dataset focuses on autism spectrum disorders and collects multimodal neuroimaging data from more than thousands of subjects. ABCD ~\cite{cai2022graph} and OpenNeuro~\cite{markiewicz2021openneuro} datasets aim to support brain development and cognitive function analysis.

\section{Challenges and Outlooks}
Despite the significant achievements obtained in MGL, there are still open research questions that require ongoing efforts in this domain. In this section, we discuss several remaining research challenges and provide our future outlooks.

\subsection{Data Imbalance across Modalities}

In real-life applications, the collected multimodal dataset may have missing or corrupted data due to various reasons, e.g., faulty sensors, database failures, and data security issues~\cite{chen2020hgmf,DBLP:conf/icml/MaZZWFZH23}. This may result in data imbalance across modalities if one or more modalities are significantly damaged and become minority modalities. Whether current MGL models can effectively handle such issues is a largely unexplored topic. As such, comprehensive studies that evaluate the reliability and trustworthiness of existing MGL models under moderate to significant modality imbalance will be highly valuable to this field. Furthermore, remediation strategies that may alleviate data imbalance issues, such as MG augmentation and data resampling, are worth in-depth investigation so as to improve the robustness of MGL models in such a context. 

\subsection{Trustworthy Multimodal Alignment}
To accurately integrate the knowledge provided by different data modalities, the process of information alignment amongst multiple modalities is a crucial step. 
We suggest that current multimodal alignment methods can be improved mainly from two aspects. First, more explicit guidance on how to match relevant information from multiple modalities is needed, e.g., locating the correct image from a document given a textual description of the image in the same document. \cite{chen2023tagging} pointed out that in many current vision-language models, the multimodal matching is done in a brute-force manner without explicit guidance. We argue that such multimodal matching is less useful for MGs, especially graph-level MGs, due to the heterogeneous topology of graphs. Therefore, developing graph structure-adaptive multimodal matching methods is needed. Second, the reliability of each modality needs to be explicitly considered during multimodal alignment. In real-life applications, due to random noises, it is common for data modalities to have different accuracies. \cite{zheng2023multi} mentioned that many multimodal classification models only focus on exploiting information aggregation, while ignoring the learning confidence of each modality. Thus, we believe that further methodological improvement from this aspect will enhance the robustness of the multimodal alignment.

\subsection{Temporal Multimodal Graph Learning}

An MG may evolve over time with changes in graph topology and node/edge features. Learning on a temporally-evolving MG is challenging, due to the dynamic interactions amongst data modalities. Performing dynamic data fusion is a much more challenging task, as demonstrated by the theoretical analysis in~\cite{DBLP:conf/icml/ZhangWZHFZP23}: Compared to the learning on static MGs, learning over temporal MGs requires the extraction of higher-order features, i.e., the spatiotemporal correlations amongst nodes. This challenges the learning capability of MGL models, especially on how to accurately align the information across different modalities on both the temporal and spatial dimensions. \cite{cai2022multimodal} proposed a multimodal learning model over temporal graphs based on continual learning, which captures the evolving graph structure using neural architecture search. This study demonstrates the potential of continual learning for this task. In the future, more advanced techniques, such as the adaptation of coninual learning, are worth being studied for this field.

\subsection{Computational Efficiency of Large-Scale MGs}
With the exponentially growing sizes of unimodal models (e.g., 110M parameters for BERT and 175B parameters for GPT3), the computational complexity of many MGL models is also rapidly growing, due to the need of processing different modalities. This poses significant challenges in the training of these models~\cite{kim2022transferring} and also the deployment of these models onto resource-constrained devices, such as laptops and mobile phones at user ends. Hence, improving the computational efficiency of MGL models will be of significant value to this field.
It has been pointed out that various existing multimodal models, e.g., DALL-E and CM3, rely on knowledge memorization, and thus, need to grow model parameters when learning large-scale datasets~\cite{DBLP:conf/icml/YasunagaAS0LLLZ23}. This calls for techniques, which guide a model to learn high-quality and generalizable features from data rather than simply memorizing and storing features. Other techniques, which deal with model sizes directly, such as model compression and distributed learning will also be valuable to study.

\section{Conclusion}
In this paper, we presented an overview of the existing techniques for representation learning on multimodal graphs. We reviewed how deep learning techniques are designed for various types of multimodal graphs to achieve effective data fusion. We discussed the characteristics of those popular designs on their effectiveness in mining the intra- and inter-modal correlations. We summarized the key applications of multimodal graph learning with crucial pointers to implementation resources. This paper provides a comprehensive discussion on the current progress and remaining challenges of this domain, which can be used as a guideline to new researchers.


\bibliographystyle{named}
\bibliography{ijcai24}

\end{document}